\documentclass[twocolumn]{revtex4-1}

\usepackage{hyperref}
\usepackage{amsfonts}
\usepackage{amsmath}
\usepackage{mathtools}
\usepackage{amssymb}
\usepackage{amsthm}
\usepackage{braket}
\usepackage[linesnumbered, ruled]{algorithm2e}
\usepackage[caption=false]{subfig}

\newcommand*{\trace}{\mathsf{Tr}} 

\begin{document}
\title{Taking gradients through experiments: LSTMs and memory proximal policy optimization for black-box quantum control}

\author{Moritz August}
\affiliation{Department of Informatics, Technical University of Munich, 85748 Garching, Germany (august@in.tum.de)}

\author{Jos\'{e} Miguel Hern\'{a}ndez-Lobato}
\affiliation{Computational and Biological Learning Lab, University of Cambridge, CB2 1PZ Cambridge, United Kingdom (jmh233@cam.ac.uk)}

\begin{abstract}
In this work we introduce the application of black-box quantum control as an interesting reinforcement learning problem to the machine learning community. 
We analyze the structure of the reinforcement learning problems arising in quantum physics and argue that agents parameterized by long short-term memory (LSTM) networks trained via stochastic policy gradients yield a general method to solving them.
In this context we introduce a variant of the proximal policy optimization (PPO) algorithm called the memory proximal policy optimization (MPPO) which is based on this analysis.
We then show how it can be applied to specific learning tasks and present results of numerical experiments showing that our method achieves state-of-the-art results for several learning tasks in quantum control with discrete and continuous control parameters.
\end{abstract}

\maketitle
\section{Introduction}
As a result of collaborative efforts by academia and industry, machine learning (ML) has in recent years led to advancements in several fields of application ranging from natural language and image processing over chemistry to medicine.
In addition to this, reinforcement learning (RL) has recently made great progress in solving challenging problems like Go or Chess~\cite{silver2017mastering, silver2017mastering1} with only small amounts of prior knowledge which was widely believed to be out of reach for the near future. 
Consequentially, RL is nowadays thought to hold promise for applications such as robotics or molecular drug design.
This success naturally raises the question of what other areas of application might benefit from the application of machine learning.

Quantum mechanics and especially quantum computing is of special interest to the machine learning community as it can not only profit from applications of state-of-the-art ML methods but is also likely to have an impact on the way ML is done in the future~\cite{biamonte2017quantum}.
This bidirectional influence sets it apart from most other applications and is a strong incentive to investigate possible uses of machine learning in the field despite the comparably steep learning curve.

One challenging and important task in the context of quantum physics is the control of quantum systems over time to implement the transition between an initial and a defined target physical state by finding good settings for a set of control parameters~\cite{nielsen2002quantum}.
This problem lies at the heart of quantum computation as performing any kind of operation on quantum bits (qubits) amounts to implementing a controlled time evolution with high accuracy in the face of noise effects induced by the environment.
Apart from the relevance to quantum computation, the analysis and understanding of the properties of quantum control problems also is an interesting research problem in its own right.
However, for a given physical system as implemented in a real experiment it is in general not possible to express all influence factors and dependencies of particles in mathematical form to perform an analytical analysis or gradient-based optimization of the control variables.
Thus, physicists have for some time been proposing automated solutions for these problems~\cite{wigley2016fast, melnikov2018active, palitta2017learning, bukov2017machine, august2017using} that are able to find good control parameter settings while being as agnostic as possible about the details of the problem in question.
Unfortunately though, these approaches are in general based on tailored solutions that do not necessarily generalize to other problems as they, e.g.\ only consider discrete variables when the underlying problem is actually continuous and are not always very sample efficient.

In this work we improve over the status quo by introducing a control method based on recurrent neural networks (RNNs) and policy gradient reinforcement learning that is generic enough to tackle every kind of quantum control problem while simultaneously allowing for the incorporation of physical domain knowledge.
More precisely, we present an improved version of the recently introduced proximal policy optimization (PPO) algorithm~\cite{schulman2017proximal} and use it to train Long Short-Term Memory (LSTM)~\cite{hochreiter1997long} networks to approximate the probability distribution of good sequences of control parameters.
We furthermore show how physical domain knowledge can be incorporated to obtain state-of-the-art results for two recently addressed control problems~\cite{bukov2017machine, august2017using}.
While our method is based on an analysis of the reinforcement problem underlying quantum control, it can also be applied to other RL problems yielding the same structure. 
Our contribution hence is threefold in that we firstly introduce the general method, secondly demonstrate how to successfully apply it to quantum control problems and thirdly, by doing so, try to stimulate a more intense exchange of ideas between quantum physics to the broader machine learning community to facilitate mutual benefit.

The rest of this work is structured as follows: in Section~\ref{q_control}, we provide a very brief introduction to quantum control, followed by a discussion and analysis of the reinforcement learning problem posed by quantum control in Section~\ref{rl_for_qc}.
Building on the analysis, we present the method in Section~\ref{algorithm} and subsequently introduce two concrete quantum control problems in Sections~\ref{quantum_memory} and \ref{ground_states} respectively.
We then present numerical results obtained by our method for these problems and compare them to those of existing solutions in Section~\ref{numerics}.
Finally, we conclude with a discussion of the work in Section~\ref{conclusion}.

\section{Quantum Control}
\label{q_control}
The time evolution of a physical system in quantum mechanics is described by the Schr\"odinger equation
\begin{align}
	ih\frac{\delta}{\delta t} \ket{\psi(t)} = H \ket{\psi(t)} 
\end{align}
where $H$ is the Hamiltonian, a complex Hermitian Matrix describing the energy of the physical system, and $h$ is Planck's constant~\cite{cohen1977quantum}. 
Hereby, $\ket{\psi}$ is the Dirac notation for a physical state which for finite dimensional systems as we treat here corresponds to a complex column vector of the same dimensionality as the Hamiltonian's. 
The conjugate transpose of a vector $\ket{\psi}$ then is denoted as $\bra{\psi}$ such that $\braket{\psi,\psi}$ denotes the inner and $\ket{\psi}\bra{\psi}$ the outer product.
The Schr\"odinger equation yields the unitary quantum time evolution 
\begin{align}
	\ket{\psi(t)} = e^{-itH/h} \ket{\psi(0)}.
\end{align}
In a discretized time setting with time steps $\Delta t$ the evolution for a total time $T$ can thus be written as
\begin{align}
	\ket{\psi(T)} = {e^{-i\Delta t H /h}}^{L} \ket{\psi(0)}
\end{align}
where we define $L = T/{\Delta t}$.
In quantum control we now assume to be able to control the time evolution by application of so-called control Hamiltonians $H_1,\cdots,H_C$, which yields the controlled time evolution
\begin{align}
	\ket{\psi(T)} = &{e^{-i\Delta t \sum_{i=1}^C c_{iL} H_i /h}}\cdots\\
&{e^{-i\Delta t \sum_{i=1}^C c_{i1} H_i /h}} \ket{\psi(0)}
\end{align}
where the $c_{it}$ are time-dependent scaling constants for the control Hamiltonians.
This formulation however assumes that we have full control over the system which due to various kinds of noise or environmental effects will not be the case.
Hence we introduce a noise or drift Hamiltonian $H_0$, which we here assume to be time independent and of constant strength, and obtain the final formulation
\begin{align}
	\ket{\psi(T)} = &{e^{-i\Delta t (H_0 + \sum_{i=1}^C c_{iL} H_i)}}\cdots\\
&{e^{-i\Delta t( H_0 + \sum_{i=1}^C c_{i1} H_i)}} \ket{\psi(0)}
\end{align}
where we set $h=1$ for convenience.

Now that we have a well-defined notion of our control problem, we need to state the actual goal that we aim to achieve.
Generally, starting from an initial state $\ket{\psi(0)}$ or the corresponding density operator $\rho(0) = \ket{\psi(0)}\bra{\psi(0)}$ we would like to obtain an evolution to target state $\ket{\psi^*}$ or $\rho^* = \ket{\psi^*}\bra{\psi^*}$.
Hence we need to define some similarity measure between the state we actually obtain after evolving for time $T$ and our ideal result.
The easiest way of doing this is simply to compute the overlap between these states by 
\begin{align} 
	S(\psi^*, \psi(T)) = \braket{\psi^*, \psi(T)}
\end{align}
or
\begin{align}
	S(\rho^*,\rho(T))= \trace {\rho^*}^{\dagger}\rho(T)
\end{align}
respectively for Hermitian operators and correspondingly only using the real part $\text{Re}(S(\rho^*,\rho(T)))$ for non-Hermitian ones~\cite{khaneja2005optimal}. 

Equipped with this metric, we can formally define the problem we would like to solve as
\begin{align}
	\max_{\{c_{it}\}} S(\rho^*, \rho(T, \{c_{it}\}).
\end{align} 
This formulation is broad enough to capture every problem from synthesizing certain quantum gates over evolving from one eigenstate of a Hamiltonian to another to storing the initial state in a quantum memory setting.

\section{Reinforcement Learning: Why and What?}
\label{rl_for_qc}
As we have seen above, solving quantum control problems amounts to determining an optimal or at least good sequence of principly continuous variables that describe the influence we exert on the system at each discrete time step.
If a rigorous mathematical description of the evolution dynamics is available, there exist methods like GRAPE~\cite{khaneja2005optimal} or CRAB~\cite{doria2011optimal, caneva2011chopped} to obtain good solutions.
However, the gap between theory and experiment also does not close in quantum mechanics and hence it is reasonable to assume that the actual dynamics of a real experiment will slightly differ from the mathematical model due to various noise effects induced by the environment.
As can for instance also be observed in robotics, these slight differences between theory/simulation and real world implementation might still have a significant impact on the optimization problem to be solved.
Additionally, it is clear that in general it is neither an interesting nor feasible task to derive a proper mathematical model for the effect of every influence factor in a real experiment~\cite{bukov2017machine}.

This shows that it is worthwhile to investigate ways of optimizing such a control problem from a black box perspective in the sense that we are agnostic about the actual time evolution dynamics of the system and can only observe the final results obtained by a chosen set of parameters.
In fact, in the absence of a mathematical model it is the only possible option to obtain information after the end of an experiment as in quantum mechanics a measurement during the experiment would in general cause the wave function to collapse and hence destroy the experiment without any way of determining what the final outcome would have been.
Hence the task we would like to solve is to find a controller or at least find a good sequence of control parameters based on the outcomes of trial runs of a given experiment, which in quantum control terminology corresponds to a closed-loop setting.

While one viable route to solving this problem would be to use classical evolutionary or hill-climbing algorithms or more advanced black-box methods such as Bayesian optimization, another interesting option is to fit a generative probabilistic model from which we can efficiently sample good sequences. 
This approach has two advantages. 
Firstly, we can iteratively update the model by fitting it to additional data we might acquire after the initial fitting phase. Doing so allows it to improve over previous results or make it adapt to changing conditions, e.g.\ a change of the noise Hamiltonian after some time. 
This is in contrast to pure optimization methods which would have to start from scratch for every problem.
Secondly, by examining the distribution over the sequence space the model has learned and inspecting the best sampled control sequences, it might be possible to gain a better understanding of the underlying dynamics of a system.

It is clear that the sequences of control parameters in a quantum control problem should not be treated as i.i.d.\ as a given choice of parameters $c_t$ at time $t$ potentially depends on all previous choices $c_1,\cdots,c_{t-1}$ and thus we have a conditional distribution $p(c_t|c_1, \cdots, c_{t-1})$. 
This kind of distribution can successfully be learned by modern RNN variants, such as LSTM or Gated Recurrent Unit (GRU) networks. 
This can for instance be seen in natural language processing (NLP) problems, which feature similar structure and where RNNs have led to breakthrough results in recent years.
Note that, with this modelling decision, we still capture the full multivariate distribution $p(c_1,\cdots,c_T)$ as by the factorization rule of probabilities it holds that
\begin{align}
p(c_1,\cdots,c_T) = \prod_{t=1}^T p(c_t|c_1,\cdots,c_{t-1}).
\end{align}

Having decided on the class of models to employ, we are left with the question of how to fit them. 
This is non-trivial as we obviously can not hope to be able to obtain gradients of real-world experiments and also can not assume to have any a priori data available. 
Hence, we must `query' the experiment to obtain tuples of sequences and results. 
Thereby we would naturally like to be as sample efficient as possible and hence have to find an intelligent way to draw samples from the experiment and learn from them.

In a recent attempt to address this problem, an evolutionary-style algorithm for training LSTMs was introduced~\cite{august2017using} that iteratively generates better data and fits the models to that data, then uses sampling from these models instead of the usual mutation operations to generate new sequences.
While the algorithm was able to find better sequences than known in theory for the considered control problem of quantum memory, it was only demonstrated for a discretized version of the problem and there is room for improvement with respect to the efficient use of sampleded sequences.

A more direct solution to this black-box optimization problem would however be if we were able to simply approximate the gradient of the error function with respect to the parameters of our model from the sampled data. Being able to obtain an approximate gradient would allow us to optimize our model in a gradient descent fashion and thus to leverage existing optimization methods mainly used in superivsed learning. Indeed, this is a typical RL scenario which is commonly referred to as \emph{policy gradient} learning. In the following, we will thus show how to solve the optimization task at hand by perceiving the problem of black-box quantum control as an RL problem and tackling it with a state-of-the-art policy gradient algorithm.
To this end, we start by analyzing the particular reinforcement learning problem posed by black-box quantum control.

As we only receive a result or measurement, from now on also referred to as reward, after having chosen a complete sequence of control parameters, we can perceive the sequence $c=(c_1,\cdots, c_T)$ as a single action of the RL agent for which it receives a reward $R(c)$. This approach most clearly reflects the envisioned closed-loop control scenario explained above. 
Modelling the sequences and their respective results in this way then implies that our Markov decision process (MDP) takes the form of a bi-partite graph consisting of a single initial state $s_0$ on the left and multiple final states $s_c$ on the right that are reached deterministically after exactly one action $c$.
The set of states $S$ of this MDP is thus given by $S = s_0 \cup \{ s_c \}$ while the set of actions $A$ corresponds to $A = \{c\}$ and the transition probabilities are defined as $P_c(s_0, s_c)=1$.
The reward $R(c)$ of an action $c$ is determined by the associated value of the error function as defined in Section~\ref{q_control}.
We assume here that two different sequences always lead to different final states of the system, which is the most challenging conceivable case as equivalence classes in the sequence space would effectively reduce the size of the search space.
This particular structure then implies that the value function simplifies to 
\begin{align}
	V(s_0) = \max_c R(c) = R(c^{opt})
\end{align}
where $c^{opt}$ is the optimal sequence and the Q-function
\begin{align}
Q(s_0,c) = R(c)
\end{align}
is in fact independent of the state and equal to the reward function $R(c)$ as each sequence $c$ is associated with exactly one final state.
Additionally, the number of actions $| \{c\} |$ and hence final states $x_c$ is \emph{at least} exponential in the number of possible values of control parameters per time step $t$ and generally infinite.
This learning setting can be perceived as a \emph{multi-armed bandit}~\cite{robbins1985some} problem but constitutes a special case as firstly we assume to be only able to perform one action, i. e.\ generate one sequence, before receiving the total reward and secondly the actions are not atomic but rather exhibit a structure we exploit for learning.

While it is true that one could derive a different formulation of the problem by considering the $c_t$ to be individual actions and using the discounted rewards of complete sequences, this approach puts more emphasis on optimal local behvaior of the agent when our goal clearly is to optimize the global performance, i.e.\ to generate the best possible sequences of control parameters.
However, for this RL problem to be solvable the compositional structure of the actions $c$ is in fact of critical importance as we will discuss now. 

In principle, the RL problem amounts to learning to choose the best out of up to infinitely many possible actions which in general clearly is unsolvable for every algorithm.
So, why can we hope to achieve something with an algorithm learning from trials in the introduced problem setting?
The main reason for this is in fact that we know that the actions the agent takes are not atomic but concatenations of multiple sub-actions which have a physical meaning. 
Nature as we perceive it seems to be governed by simple underlying rules (or complex rules that are at least approximated very well by simple ones) which allows us to capture them with mathematical expressions. 
This in turn implies that there is much structure to be found in Nature and hence it is reasonable to assume that likewise the desirable actions in our learning problem share certain patterns which can be discovered.
More precisely, we conjecture that solving the particular problems we are tackling in this work requires less abstract conceptual inference, which would still be out of reach for todays machine learning models, and more recognition of patterns in large sets of trials, i.e.\ control sequences, and hence in fact lends itself to treatment via machine learning and especially contemporary RNN models.
Some empirical evidence for the validity of this conjecture has recently been provided for the problem of quantum memory~\cite{august2017using} and for a problem related to quantum control, the design of quantum experiments~\cite{melnikov2018active}.

\section{The Learning Algorithm}
\label{algorithm}
Having discussed the modelling of the control sequences and the RL problem, we will now introduce the actual learning algorithm we employ.
As we have seen above, we can not perform direct optimization of $R(c)$ as we cannot access $\nabla R(c)$.
However, it has long been known that it is possible to approximate $\nabla_{\Theta} \mathbf{E}_c[R(c)]$ since
\begin{align}
\nabla_{\Theta} \mathbf{E}_c[R(c)] = \mathbf{E}_c[\nabla \ln p_{\Theta}(c) R(c)]
\end{align}
where $\mathbf{E}_c$ is the expectation over the sequence space and $p_{\Theta}(c)$ is the stochastic policy of the agent parameterized by the weight vector $\Theta$, which in this work corresponds to an RNN\@.
This insight is known as the likelihood ratio or REINFORCE~\cite{williams1992simple} trick and constitutes the basis of the policy gradient approach to reinforcement learning.
From the physics point of view, the trick allows us to take the gradient of the average outcome of a given experiment with respect to the parameters of our stochastic controller and perform gradient-based optimization while being agnostic about the mechanisms behind the experiment, i.e.\ model-free. In a sense we thus have a way of taking a gradient through an experiment without the necessity to mathematically model every variable of influence and their interplay.
From a different perspective, this approach simply corresponds to maximizing the likelihood of sequences that are weighted by their results, such that the agent has a higher incentive to maximize the likelihood of good sequences.
The approach can be refined by replacing the weighting by the pure reward $R(c)$ with an approximation of the advantage $A(s,c) = Q(s,c) - V(s)$.
This has been shown to improve the convergence significantly and especially for continuous control problems, policy gradient methods outperform Q-learning algorithms~\cite{schulman2017proximal}.

Despite such improvements, policy gradient approaches still suffer from slow convergence or catastrophicly large updates, which has led to the development of improvements such as trust region policy optimization~\cite{schulman2015trust} (TRPO).
These methods however make use of second-order information such as inverses of the Hessian or Fisher information matrix and hence are very difficult to apply in large parameter spaces which are common in the deep learning regime.
The underlying idea of such improvements thereby is limiting the magnitude of updates to $\Theta$ by imposing constraints on the difference between $p_{\Theta}$ and $p_{\Theta_{new}}$ in order to prevent catastrophic jumps out of optima and achieve a better convergence behvaior.

In an effort to strike a balance between ease of application and leveraging the insights behind TRPO, recently a novel policy gradient scheme called proximal policy optimization~\cite{schulman2017proximal} (PPO) was introduced.
One main novelty hereby lies in the introduced loss, which is for a general RL scenario given by
\begin{align}
L^{CLIP}(\Theta) = &\mathbf{E}_t[\min(r_t(\Theta)A_t, \\
&\text{clip}(r_t(\Theta),1-\epsilon, 1+\epsilon)A_t)]
\end{align}
where $\mathbf{E}_t$ and $A_t$ are the expectation over time steps and the advantage at time $t$ respectively, which both need to be approximated.
The term $r_t$ is defined as the ratio of likelihoods
\begin{align}
r_t(\Theta) = \frac{p_{\Theta}(c_t|s_t)}{p_{\Theta_{old}}(c_t|s_t)}
\end{align}
of actions $c_t$ in states $s_t$ in our notation and we define $\text{clip}(a,b,c)=\text{min}(\text{max}(a,b),c)$.
The distribution $p_{\Theta}(c_t|s_t)$ is a stochastic policy depending on parameters $\Theta$.
Note that this generic formulation assumes multiple actions $c_t$ per episode and thus does not yet apply to the learning scenario discussed here.

The objective function poses a lower bound on the improvement induced by an update and hence establishes a trust region around $\Theta_{old}$.
The hyperparameter $\epsilon$ controls the maximal improvement and thus the size of the trust region.

Now, the basic algorithm is defined as follows:
\begin{enumerate}
	\item Obtain new set of trajectories, i.e.\ sequences, $C$, by sequentially sampling from $p_{\Theta}(c_t|s_t)$.
	\item Optimize $L^{CLIP}$ over $C$ for $K$ iterations.
	\item Set $\Theta_{old} = \Theta$.
	\item Repeat until convergence.
\end{enumerate}
Note that there exists a straight-forward generalization to the case of multiple agents but as we can not reasonably assume in our application to have access to multiple identical experiments, we only consider the case of one agent here. The algorithm was shown to achieve state-of-the-art performance for several discrete and contiouous control tasks, which makes it ideally suited for the problems tackled in this work. 
However, we will now introduce a few improvements tailored to our specific reinforcement learning problem as defined in the previous section which we will for the sake of brevity from now on refer to as memory proximal policy optimization (MPPO).

Since in our problem we only consider episodes consisting of one action $c$, the objective becomes\begin{align}
L_1^{CLIP}(\Theta) = &\mathbf{E}_c[\min(r(\Theta)A,\\ 
&\text{clip}(r(\Theta),1-\epsilon, 1+\epsilon)A)]
\end{align}
with
\begin{align}
r(\Theta, c) = \frac{p_{\Theta}(c)}{p_{\Theta_{old}}(c)}
\end{align}
and $p_{\Theta}(c)$ being parameterized by an LSTM, as discussed above. $A$ again denotes the advantage function.
We have omitted the dependence on $c$ in $L_1$ for the sake of clarity.
Since we know that in our problem setting it holds that $Q(c,s)=R(c)$, the advantage function becomes 
\begin{align}
A(c) = R(c) - V(c).
\end{align}
It is worth noting that this implies that in our scenario there is no need to approximate the Q-function as we can access it directly. 
In fact approximating the Q-function and hence $R(c)$ would be equivalent to solving the optimization problem as we could use the approximator to optimize over its input space to find good sequences. 
The quality of the approximation of $A(c)$ consequentially only depends on the approximation of $V(c)$. 
While there exist many sophisticated ways of approximating the value function~\cite{schulman2017proximal, mnih2016asynchronous} in our case the optimal approximation is given by 
\begin{align}
\hat{V}(c) = R(c^*)
\end{align}
where $c^*$ is the best sequence we have encountered so far. Since we do not know the best sequence and its corresponding reward (at best we know an upper bound), the reward of the best sequence found so far is the closest approximation we can make. 
The optimal approximation of the advantage $A(c)$ hence is given by
\begin{align}
\hat{A}(c) = R(c) - R(c^*).
\end{align}
Since we need to store $c^*$ to compute the advantage approximation and are generally interested in keeping the best solution, it is a natural idea to equipping the agent with a memory $M$ of the best sequences found so far. 
We can then formulate a memory-enhanced version of the PPO algorithm:
\begin{enumerate}
	\item Obtain new set of trajectories, i.e.\ sequences, $C$, by sampling from $p_{\Theta}(c)$.
	\item Update the memory of best sequences $M$
	\item Optimize $L_1^{CLIP}$ over $C \cup M$ for $K$ iterations.
	\item Set $\Theta_{old} = \Theta$.
	\item Repeat until convergence
\end{enumerate}
The memory sequences are treated as newly sampled sequences such that their weighting always is performed with respect to the current values of $\Theta_{old}$ and $\Theta$.
This ensures compatibility with the policy gradient framework while the access to the best actions discovered so far leads to a better convergence behavior as we will see later.
Note that, under the previously introduced assumption, the best sequences share common structural properties. Maximizing the expected reward over all sequences $\mathbf{E}_c[R(c)]$ is thus equivalent to maximizing the expected reward over the sequences in the memory $\mathbf{E}_{c \in M}[R(c)]$ which ensures relevance and stability of the updates computed over $M$.
This memory scheme furthermore is different from experience replay in Q-learning~\cite{mnih2015human} as only the best sequences are kept and reintroduced to the agent.
The relation between $|C|$ and $|M|$ thereby is a new hyperparameter of the algorithm affecting the exploration-exploitation dynamics of the learning process.

Another factor that has a significant impact on the exploration behavior is the value of the scaling or variance parameter of the probability distributions employed in continuous control tasks, such as for instance the standard deviation $\sigma$ of the univariate normal distribution or the covariance matrix $\Sigma$ in the multivariate case. 
It is clear that a large variance induces more exploration while a small variance corresponds to a more exploitation-oriented behavior.
Over the course of training an agent to find a good policy it is hence reasonable to start with a larger variance and reduce it during the optimization until it reaches a defined minimal value. 
However, while the agent usually learns to predict the mean of the given distribution, the variance parameter is currently often treated as fixed or follows a predefined decay schedule which does not account for the randomness in the training process.
Utilizing the sequence memory, we propose an improvement by introducing a dynamical adaptation scheme for the variance parameters depending on the improvement of the memory $M$.
More concretely, we propose to maintain a window $W_i$ of the relative improvements of the average rewards in memory
\begin{align}
	W_i = \left[ \frac{\overline{ R(M_{i-l+1})} -\overline{R(M_{i-l})}}{\overline{ R(M_{i-l})}},\cdots,\frac{\overline{(M_i)} -\overline{R(M_{i-1})}}{\overline{R(M_{i-1})}} \right]
\end{align}
where $\overline{R(M_i)}$ denotes the average reward over the memory in iteration $i$ of the optimization and $l$ is the window length. 
At every $l$-th step in the optimization, we then compute a change parameter
\begin{align}
	\alpha_t = 1 + \frac{\overline{W_{t-l}} - \overline{W_t}}{\overline{W_{t-l}}}
\end{align}
with $\overline{W_t}$ being the window average and multiply (possibly clipped) the variance parameters by it.
Note that we assume here monotonic improvement of $M$ and $R \in [0,1]$. 
This scheme thus poses a dynamic adaptation of the variance parameters based on second-order information of the improvement of the average reward of $M$. 
It follows the intuition that if the improvement slows down, a decrease of the variance gives the agent more control over the sampled actions and allows for a more exploitation-oriented behavior.
On the other side, when the improvement accelerates, it appears reasonable to prevent too greedy a behavior by increasing the uncertainty in the predicted actions. The same scheme can furthermore also be applied to parameters such as $\epsilon$, which plays a similar role to the variance.

In conclusion, extending the PPO training with a memory of the best perceived actions prevents good solutions of the control problem to be lost, gives the agent access to the best available advantage estimate, improves convergence and allows to dynamically scale the variance parameters of respective distributions from which actions are sampled.
While we introduce this variant of the PPO algorithm for our specific application, we believe that it would generalize to other applications of reinforcement learning.

\section{Applying the Method}
In this section, we will now introduce two quantum control scenarios that were recently explored via machine learning~\cite{bukov2017machine, august2017using}. 
We show how one can apply our method to tackle some interesting learning tasks arising in these control settings by leveraging physical domain knowledge.

\subsection{Quantum Memory}
\label{quantum_memory}
One particular instance of a quantum control problem is the problem of storing the state of a qubit, i.e.\ a two-level system used in quantum computation. This is, next to quantum error correction, a very relevant problem in quantum computation.
Here we assume that our qubit is embedded in some environment, called the bath, such that the complete system lives in the Hilbert space
\begin{align*}
	\mathcal{H} = \mathcal{H}_S \otimes \mathcal{H}_B
\end{align*}
with the subsripts $S$ and $B$ denoting the space of the system and bath respectively. 
If we let this system evolve freely, decoherence effects will over time destroy the state of the qubit.
Hence the question is how we can intervene to prevent the loss of the state in the presence of the environment or, for computer scientific purposes, the noise where we assume to have control over the qubit only. From a quantum computing perspective, we would like to implement a gate that performs the identity function over a finite time interval.

Qubit states are commonly represented as points on the Bloch sphere~\cite{nielsen2002quantum} and the effect of the environment on the qubit can in this picture be perceived as some rotation that drives the qubit away from its original position.
To counter this problem we must hence determine a good rotation at each time step such that we negate the effect of the environment.
So, our goal is to dynamically decouple the qubit from its bath by performing these rotations.
The rotation of a qubit is defined as
\begin{align*}
	R_n(\alpha) = e^{-i\frac{\alpha}{2}n\mathbf{\sigma}}
\end{align*}
with $n$ being a unit vector specifying the rotation axis, $\alpha$ denoting the rotation angle and $\mathbf{\sigma}$ the `vector' of the stacked Pauli matrices $\sigma_{\{x,y,z\}}$~\cite{sakurai1995modern}.
Thus our controlled time evolution operator per time step $t$ becomes
\begin{align*}
	U(n_t, \alpha_t) = e^{-i \Delta t ( H_0 + \frac{\alpha_t }{2\Delta t}n_t\mathbf{\sigma} \otimes I_B)},
\end{align*}
expressing that we only apply the rotation to the qubit, but not the bath. 
The noise Hamiltonian $H_0$ here reflects the effect of the bath on the qubit and $I_B$ simply denotes the identity of size of the dimensionality of $\mathcal{H}_B$ such that the Kronecker product yields a matrix of equal size to $H_0$.

One possible metric to quantify how well we were able to preserve the qubit's state is
\begin{align*}
	D(U,I) = \sqrt{1 - \frac{1}{d_S d_B} \Vert \trace_S(U)\Vert_{\trace}}
\end{align*}
with $U$ denoting the total evolution operator, $I$ the identity and $\trace_S$ is the partial trace over the system~\cite{quiroz2013optimized}.
$\Vert U \Vert_{\trace} = \trace \sqrt{U^{\dagger}U}$ is the trace or nuclear norm.
This distance measure is minimized by the ideal case $U = I_S \otimes U_B$ with an arbitrary unitary $U_B$ acting on the bath.
Thus, the problem we would like to solve is a special instance of quantum control and can be formulated as
\begin{align*}
	\min_{\{(n_t, \alpha_t)\}} D(U(\{(n_t, \alpha_t)\}, I).
\end{align*}
Having introduced the quantum memory scenario, we now turn to a description of possible reinforcement learning tasks in this context.
We present three different formulations of the setting which we will in the following refer to as the discrete, semi-continuous and continuous case. 
These formulations differ in the parametrization of the rotation $R_n(\alpha)$ that is to be performed at each time step.
\begin{description}
	\item[Discrete case] It is known from analytical derivations that the Pauli matrices $\sigma_{\{0, x,y,z\}}$ give rise to optimal sequences under certain ideal conditions~\cite{viola1999dynamical, souza2011robust}, where at each time step exactly one of the rotations $R_{\{0,x,y,z\}}=e^{-i \frac{\pi}{2} \sigma_{\{0,x,y,z\}}}$ is performed.
$\sigma_0$ hereby denotes the identity.
Hence, in the simplest formulation we can define the problem as choosing one of the four Pauli matrices at each time step.
This formulation then leads to a sequence space $S$ of size $|S|=4^T$ being exponential in the sequence length $T$.
This is the formulation which was also used in recent work on quantum memory~\cite{august2017using}.

\item[Semi-continuous case]
	While the class of sequences introduced above is provably ideal under certain conditions, one might be interested in allowing the agent more freedom to facilitate its adaption to more adverse conditions.
	This can in a first step be achieved by allowing the agent full control over the rotation angle while keeping the discrete formulation for the rotation axis.
	That means that at each time step, the agent will have to choose a rotation axis from $\sigma_{\{0,x,y,z\}}$ as before, but now must also predict the rotation angle $\alpha \in [0,2\pi]$.
	As $\alpha$ can take infinitely many values, this formulation of the problem now yields a sequence space $S$ of inifinite size, making it much harder from a reinforcement learning perspective.
	To lighten this burden we can make use of the fact that we know that in principle a rotation around $\pi$ is ideal.
	Thus, we will interpret the output of the agent as the deviation from $\pi$ $\Delta \alpha \in [-\pi, \pi]$.
	This should facilitate learning progress even in the early training phase.

	\item[Continuous case] Finally, we can of course also allow the agent full control over both the rotation angle and axis.
		This formulation of the problem requires the agent to predict a unit vector $n \in \mathbb{R}^3$ and a corresponding rotation angle $\alpha$ for each time step.
		It is clear that without any prior knowledge it will be very difficult for the agent to identify the `right corner' of this infinite sequence space.
		We hence propose to again leverage the knowledge about Pauli rotations being a good standard choice by having the agent predict a Pauli rotation together with the deviation in $n$ and $\alpha$.
		While for $\alpha$ we have already seen how this can be easily achieved, $n$ requires slightly more insight. 
		As is customary in quantum physics, every state of a two-dimensional particle $\ket{\psi}$ can be represented by choosing two angles $\theta \in [0,\pi]$ and $\phi \in [0, 2\pi]$, yielding the three-dimensional real unit Bloch vector
\[b =
\begin{pmatrix}
	\sin \theta \cos \phi \\
	\sin \theta \sin \phi \\
	\cos \theta
\end{pmatrix}.
\]
We can hence use this formulation to parameterize $n$ by $\theta$ and $\phi$. 
It is easy to see that the Pauli rotations correspond to the unit vectors that equal a one-hot encoding of the Pauli matrices such that we obtain the following identities
\begin{align*}
	\theta_x &= \theta_y = \phi_y = \frac{\pi}{2} \,\, \text{and} \\
	\phi_x &= \phi_z = \theta_z = 0
 \end{align*}
 with periodicity in $\pi$.
We can now leverage this knowledge by translating the Pauli rotation axis chosen by the agent into its Bloch expression and requring it to predict the deviations $\Delta \theta$ and $\Delta \phi$.
In this way the agent has access to the full axis space.
As with the rotation angle, this formulation has the effect that the agent starts learning from a reasonable baseline.
\end{description}

\subsection{Ground state transitions}
\label{ground_states}
Another scenario that was recently addressed in an anlysis of the characteristics of the optimization problem behind controlling systems out of equilibrium~\cite{bukov2017machine} is the transition between ground states of different Hamiltonians.
The considered class of Hamiltonians was thereby defined to be the class of Ising Hamiltonians given by
\begin{align*}
H(J,g,h) &= J \sum_{i=1}^{L-1} I^{\otimes i-1} \otimes \sigma_x \otimes \sigma_x \otimes I^{\otimes L-(i+1)} \\
&+ g \sum_{i=1}^{L} I^{\otimes i-1} \otimes \sigma_z \otimes I^{\otimes L - i} \\
&+ h \sum_{i=1}^{L} I^{\otimes i-1} \otimes \sigma_x \otimes I^{\otimes L - i}
\end{align*} 
where the $\sigma_{\{x,y,z\}}$ again denote the Pauli matrices and $L$ specifies the number of particles. 
In this setting we furthermore set $J=g=-1$, leaving $h$ as the only free parameter specifing the strength of the magnetic field represented by $\sigma_x$.
From a mathematical perspective, the ground state $\ket{E_{min}(h)}$ of a given Hamiltonian $H(h)$ is then defined as the eigenvector of $H(h)$ corresponding to its lowest eigenvalue.

In the considered scenario we now choose the initial and target states to be $\ket{\psi_i}=\ket{E_{min}(h_i)}$ and $\ket{\psi^*}=\ket{E_{min}(h^*)}$ respectively where $h_i \neq h^*$ are particular choices of $h$.
The controlled time evolution operator is then simply defined to be the one generated by $H(h)$ as given by
\[
U(h_t) = e^{-i \Delta t /h H(h_t)}
\]
where we assume $h_t$ to be time dependent.
The closeness between the state resulting from the controlled time evolution $\ket{\psi(T)}$ and the target state $\ket{\psi^*}$ is measured by their squared overlap
\[
S_2(\psi^*, \psi(T)) = |\braket{\psi^*, \psi(T)}|^2,
\]
similar to what was shown in Section~\ref{q_control}. 
We thus obtain the optimization problem formulation
\[
\max_{\{h_t\}} S_2(\psi^*, \psi(\{h_t\}))
\]

representing the quantum control optimization problem.

Next, we will introduce some RL tasks arising in this control scenario.
Similarly to the the taxonomy introduced above, we will thereby distinguish between a discrete, a continuous and a constrained case.
These cases correspond to different domains of possible values for the time dependent field strengths $h_t$. 
All of them however have in common that we assume a maximal magnitude $h_{max}$ of the field strength such that $h_t \in [-h_{max}, h_{max}]$ holds.
This is simply done to reflect the fact that in real experiments infinite field strengths are impossible to achieve.

\begin{description}
			\item[Discrete case] Knowing that the potentially continuous domain of our control parameter $h_t$ is upper and lower bounded by $\pm h_{max}$, we can apply Pontryagin's principle to limit ourselves to actions $s_t \in \{-h_{max}, h_{max}\}$. 
				We thus obtain a reinforcement learning problem where at each point in time the agent has to make a binary decision. 
				While this is arguably the easiest conceivable scenario, the sequence space still is of size $|S|=2^T$.

			\item[Continuous case] Although we know from theory that optimal sequences will comprise only extremal values of the control parameter $h_t$, it is still interesting to examine if the agent is able to discover this rule by itself.
				In this case we hence allow the agent to freely choose $h_t \in [-h_{max}, h_{max}]$ which again presents us with a sequence space of infinite size.
				Following our reasoning from the continuous quantum memory case, we cast the problem as learning the deviation $\Delta h \in [0, h_{max}]$ from $\pm h_{max}$.
				Hence, for each time $t$ the agent must predict the deviation $\Delta h$ and decide to which of the two extremal values the deviation should be applied.
				This formulation clearly allows the agent to predict any value in $[-h_{max}, h_{max}]$.

			\item[Constrained case] In the continuous case as defined above, we know that the agent should ideally learn to predict deviations of 0 to achieve sequences with extremal values of $h_t$.
					We can thus try to make the problem more challenging by imposing an upper bound $B < T|h_{max}|$ on $\sum_t |h_t|$, representing an upper limit of the total field strength.
					Imposing such a bound is not an artificial problem as it could for instance be used to model energy constraints in real experiments.
					This constraint can easily be realized by defining the reward of a sequence $s$ to be
					\[
					R(s) = 
					\begin{cases}
							S_2(\psi^*, \psi(s)) \; \text{if} \; \sum_t |h_t| \leq B \\
								0 \; \text{else}.
									
							\end{cases}
							\]
							This constraint requires the agent to learn how to distribute a global budget over a given sequence where it can maximally allocate an absolute field strength of $|h_{max}|$ to each action $s_t$.
							As it is not clear which values are optimal in principle for a given bound $B$, instead of a deviation we here let the agent directly predict the field strength $h_t$.

					\end{description}

\section{Results}
\label{numerics}
\begin{table}
\caption{The best values of $D(U,I)$ found by or method for the discrete, semi-continuous and continuous quantum memory learning tasks together with baseline results. The reference values were taken from~\cite{august2017using} and computed with the corresponding algorithm for $T=0.512$ and $\Delta t=0.002$. Lower values are better.}
	\begin{center}
		\begin{tabular}{c *{4}{c} }
			  & \multicolumn {2}{c}{$\Delta t =0.002$} & \multicolumn {2}{c}{$\Delta t =0.004$}\\ \hline
 			  $T=$ & $0.064$ & $0.512$ & $0.256$ & $0.512$\\ \hline
			  Ref. & $7\cdot 10^{-5}$ & $2\cdot 10^{-4}$ & $4\cdot 10^{-4}$ & $8\cdot 10^{-4}$ \\ \hline
			  Disc. & $7\cdot 10^{-5}$ & $2\cdot 10^{-4}$ & $4\cdot 10^{-4}$ & $8\cdot 10^{-4}$ \\ \hline
			Semi-Cont. & $6\cdot 10^{-5}$ & $2\cdot 10^{-4}$ & $4\cdot 10^{-4}$ & $8\cdot 10^{-4}$ \\ \hline
			Cont. & $6\cdot 10^{-5}$& $2\cdot 10^{-4}$ & $4\cdot 10^{-4}$ & $7\cdot 10^{-4}$ \\ \hline
		\end{tabular}
	\end{center}
\label{tab_qm_0}
\end{table} 

\begin{figure}[t]
	\centering
	\includegraphics[width=0.4\textwidth]{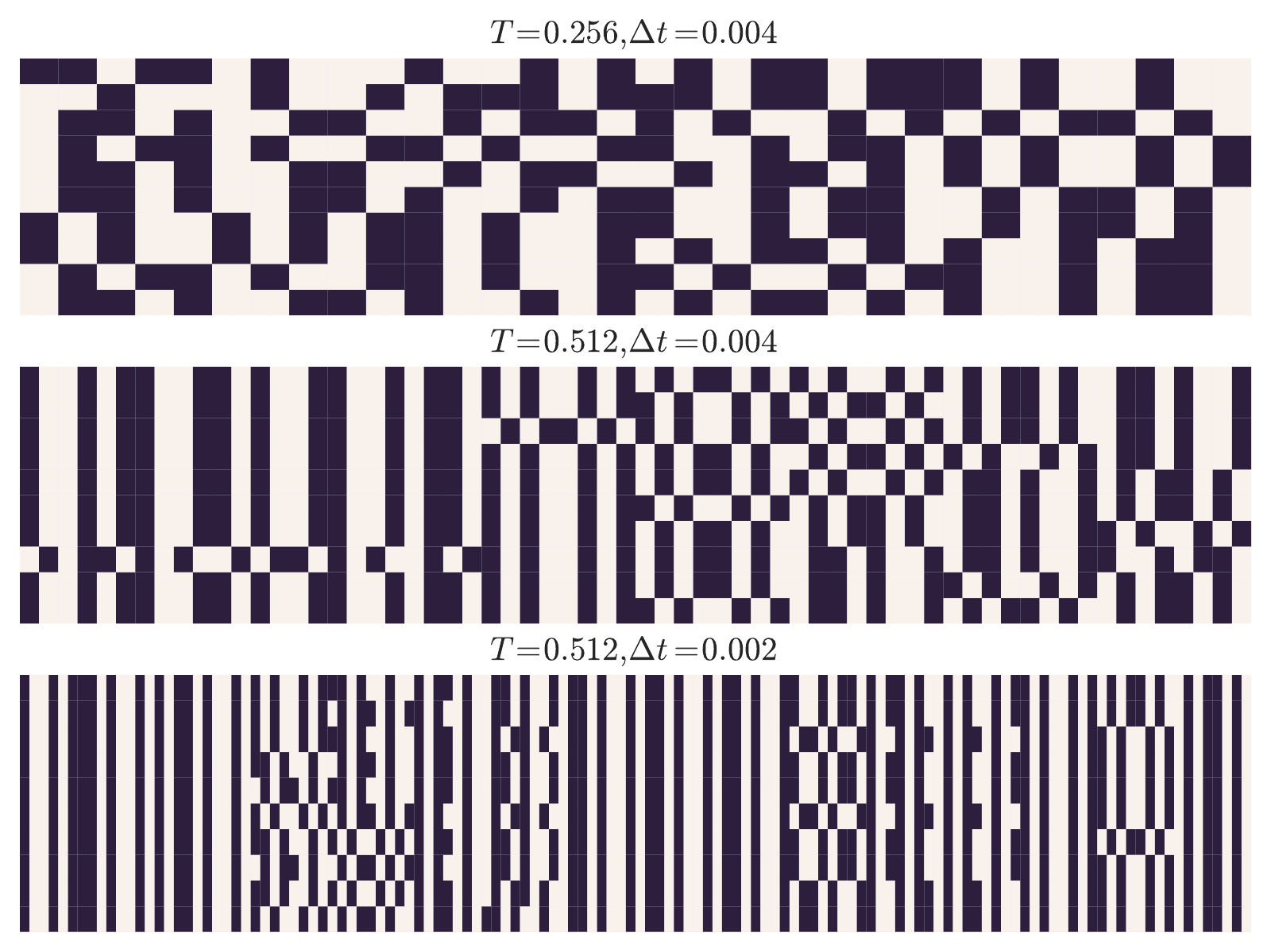}
	\caption{The best 10 sequences found for the discrete learning problem with varying parameters of $T$ and $\Delta t$. It is clearly visible how the best sequences for each setting share common structural properties and also exhibit recurring patterns making them amenable to machine learning models.}
\label{fig:qm_disc_seqs}
\end{figure}

\begin{figure}[t]
	\centering
	\includegraphics[width=0.4\textwidth]{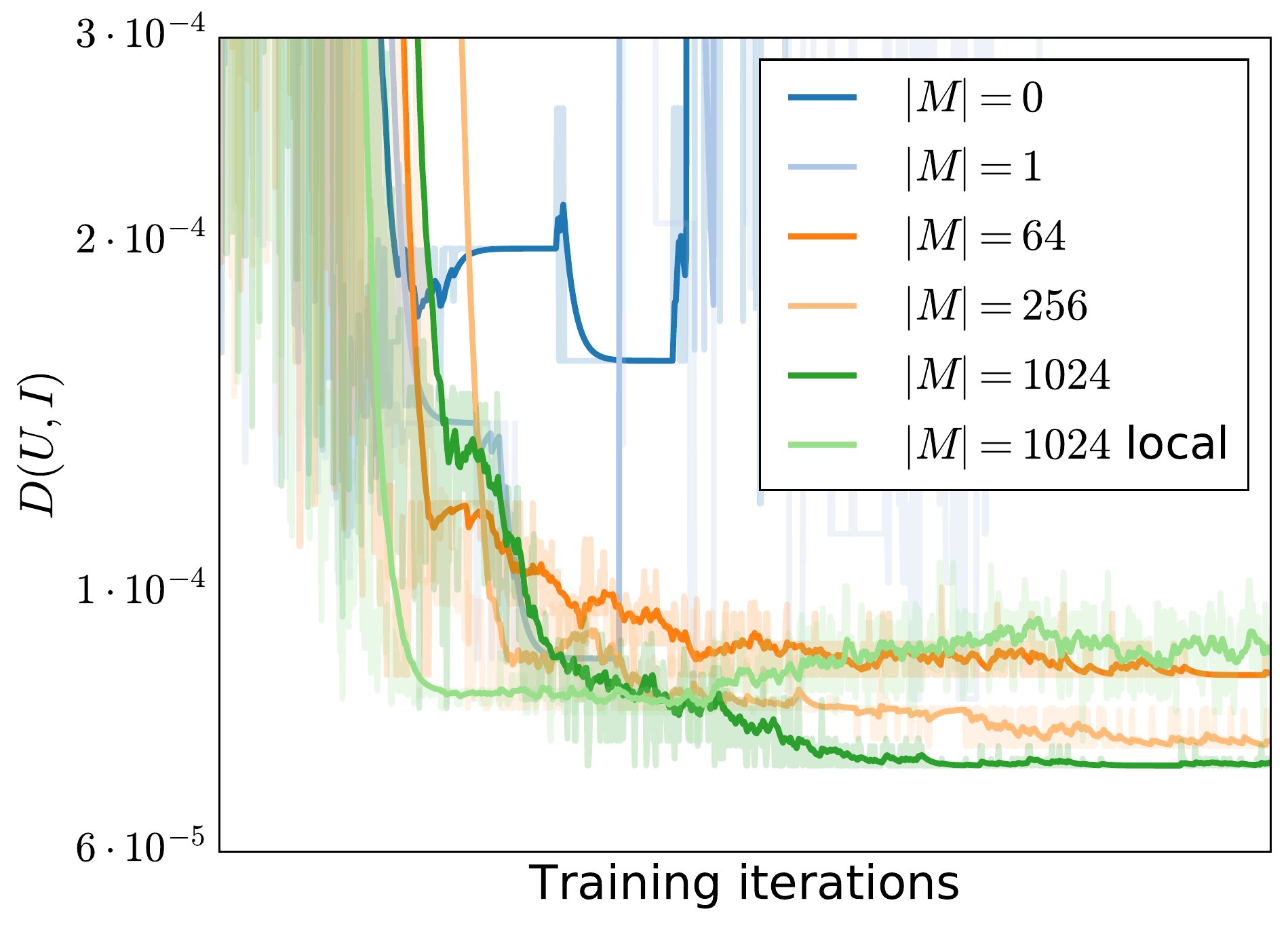}
	\caption{A comparison of the convergence behavior of the best results sampled per iteration for different sizes of the memory, no memory and a memory with the $L^{CLIP}$ loss applied to the invididual $c_t$ for $T=0.064$ and $\Delta t=0.002$. The convergence becomes more stable with larger memory and updates based on the entire sequences lead to convergence to better sequences.}
\label{fig:mem_eval}
\end{figure}

In this section we will now present numerical results for the two application scenarios presented above to illustrate the validity of our method and the usefulness of the MPPO algorithm. As we did not have at our disposal real physical experiments implementing these scenarios, the results presented in the following are based numerical simulations.
\subsection{Quantum Memory} 
For the quantum memory scenario, we investigate the performance of our algorithm for different lengths of the discrete time step, total evolution times and across the three formulations of the problem described above. 
More concretely, we explore the method's behavior for a discrete time evolution with $\Delta t \in\{0.002, 0.004\}$, $T \in \{0.064, 0.256, 0.512, 1.024\}$ and a physical system consisting of one memory qubit coupled to a bath of four qubits with up to three-body interactions to allow for a comparison with the baseline results~\cite{august2017using}.
We refer the interested reader to this article for a more precise description of the physical setup. While we ultimately would like to optimize $D(U(c),I)$ as defined above, we used $1-D(U(c),I)$ as a reward signal to obtain an $R(c) \in [0,1]$. We furthermore shifted the reward such that a uniformly random policy obtains zero reward on average.

As the three learning tasks introduced for this scenario differ in their action domains, we need to use a different probabilistic modelling for each setting.
For the discrete case, we simply model each element $c_t$ of a sequence $c$ by a categorical distribution such that we have 
\[
p(c) = \prod_t Cat(c_t \in \{I,X,Y,Z\}|\{p_{I,t}, p_{X,t}, p_{Y,t}, p_{Z,t}\})
\]
for a complete sequence $c$.
In the semi-continuous case we employ a mixture-of-Gaussians distribution which yields
\[
p(c) = \prod_t \sum_{i \in \{I,X,Y,Z\}} p_{i,t} \mathcal{N}(c_t=\Delta \alpha| \mu_{i,t}, \sigma_t). 
\]
This can easily be generalized to the continuous case via a multivariate mixture-of-Gaussians distribution with diagonal covariance matrix such that we obtain
\[
p(c) = \prod_t \sum_{i \in \{I,X,Y,Z\}} p_{i,t} \mathcal{N}(c_t=\{\Delta \alpha,\Delta \theta,\Delta \phi\}| \boldsymbol{\mu}_{i,t}, \sigma_tI). 
\]
Note that we have omitted here the dependence on the weights $\Theta$ for the sake of brevity.
As discussed in Section~\ref{rl_for_qc}, we use an LSTM to parameterize these probability densities. 
More concretely, we use a two-layer LSTM and use its output as input to a softmax layer to predict the $p_{i,t}$. 
From this output state and the relevant parts of the output from the previous time step we also predict the $\mu_i$ for $\Delta \alpha$ in the semi-continuous case and analogously for $\Delta \theta$ and $\Delta \phi$ in the continuous case. 
For every deviation output we train an individual output unit for each discrete rotation. For the semi-continuous and continuous tasks, we scale the standard deviation $\sigma_t$ and PPO parameter $\epsilon$ over the course of the optimization using our introduced adaption scheme with a window size of 10 and optimize the loss function with the Adam optimizer~\cite{kingma2014adam}.

The scores $D(U(c),I)$ of the best sequences found in our numerical experiments are listed in Table~\ref{tab_qm_0}. 
They clearly show that our method is able to achieve the same or slightly better results as the baseline algorithm from~\cite{august2017using} for all considered settings and learning tasks. 
For the semi-continuous case, we observe that for the setting involving the shortest sequences slightly better sequences than in the discrete case can be found. For longer sequences the performance is on par with the discrete sequences.
The same in principle holds for the continuous case with the exception of the results for $T=0.512$ and $\Delta t=0.004$ being slightly better then for the other two cases.
Overall we can conclude that our method finds sequences several orders of magnitude better than those a random policy generates, which are generally in the interval $[0.1, 0.5]$, showing that in all cases LSTMs trained by the MPPO algorithm seem to perform quite well.
We can also see that the discrete sequences pose a strong baseline that is hard to beat even with a fully continuous approach and in fact we observed the predicted deviations to converge to very small values. The results furthermore support the conjecture that good sequences share common structure and local patterns that can be learned which is also illustrated in Figure~\ref{fig:qm_disc_seqs}. 
Here, the best 10 sequences found during the training process in the discrete case for three different settings are shown, illustrating the high degree of structure that the best sequences exhibit. 
The structural similarities become more apparent with growing sequence length. Interestingly, in all cases the best sequences only make use of two of the four Pauli rotations and less surprisingly never use the identity `rotation'.
In Figure~\ref{fig:mem_eval} we show the effect of different sizes of the memory $M$ on the convergence of the best sequences in the discrete case for otherwise constant optimization parameters. 
As can be seen, when not using a memory or only storing the best sequence, the optimization diverges. 
For larger sizes of the memory, the algorithm converges to better and better sequences, arriving at the best sequence found for this setting with a memory of 1024 sequences. 
We also compared the performance of our algorithm to updates computed not over complete sequences but over the single control parameters $c_t$ as done in the PPO algorithm for $|M|=1024$.
While also the latter performs well, only the former converges to the best sequence.

\begin{table}
\caption{The best values of $S_2$ obtained by our method for the discrete, continuous and constrained ground state transition learning problems with reference values taken from~\cite{bukov2017machine}. Higher values are better.}
	\begin{center}
		\begin{tabular}{c *{4}{c} }
			  & $T=0.5$ & $T=1$ & $T=3$ \\ \hline
			Ref. ($L=1$) & $0.331$ & $0.576$ & $1$ \\ \hline
			Disc. ($L=1$) & $0.331$ & $0.576$ & $1$ \\ \hline
			Cont. ($L=1$) & $0.331$ & $0.576$ & $1$ \\ \hline
			Disc. ($L=5$) & $0.57$ & $0.767$ & $1$ \\ \hline
			Cont. ($L=5$) & $0.57$ & $0.768$ & $1$ \\ \hline
			Const. ($B=20$) & $0.313$ & $-$ & $-$ \\ \hline
			Const. ($B=30$) & $0.322$ & $-$ & $-$ \\ \hline
			Const. ($B=40$) & $-$ & $0.572$ & $-$ \\ \hline
			Const. ($B=50$) & $-$ & $0.577$ & $-$ \\ \hline
			Const. ($B=60$) & $-$ & $0.577$ & $-$ \\ \hline
			Const. ($B=120$) & $-$ & $-$ & $1$ \\ \hline
			Const. ($B=140$) & $-$ & $-$ & $1$ \\ \hline
			Const. ($B=160$) & $-$ & $-$ & $1$ \\ \hline
		\end{tabular}
	\end{center}
\label{tab_gs_0}
\end{table} 


\begin{figure}[h]
	\centering
	\includegraphics[width=0.4\textwidth]{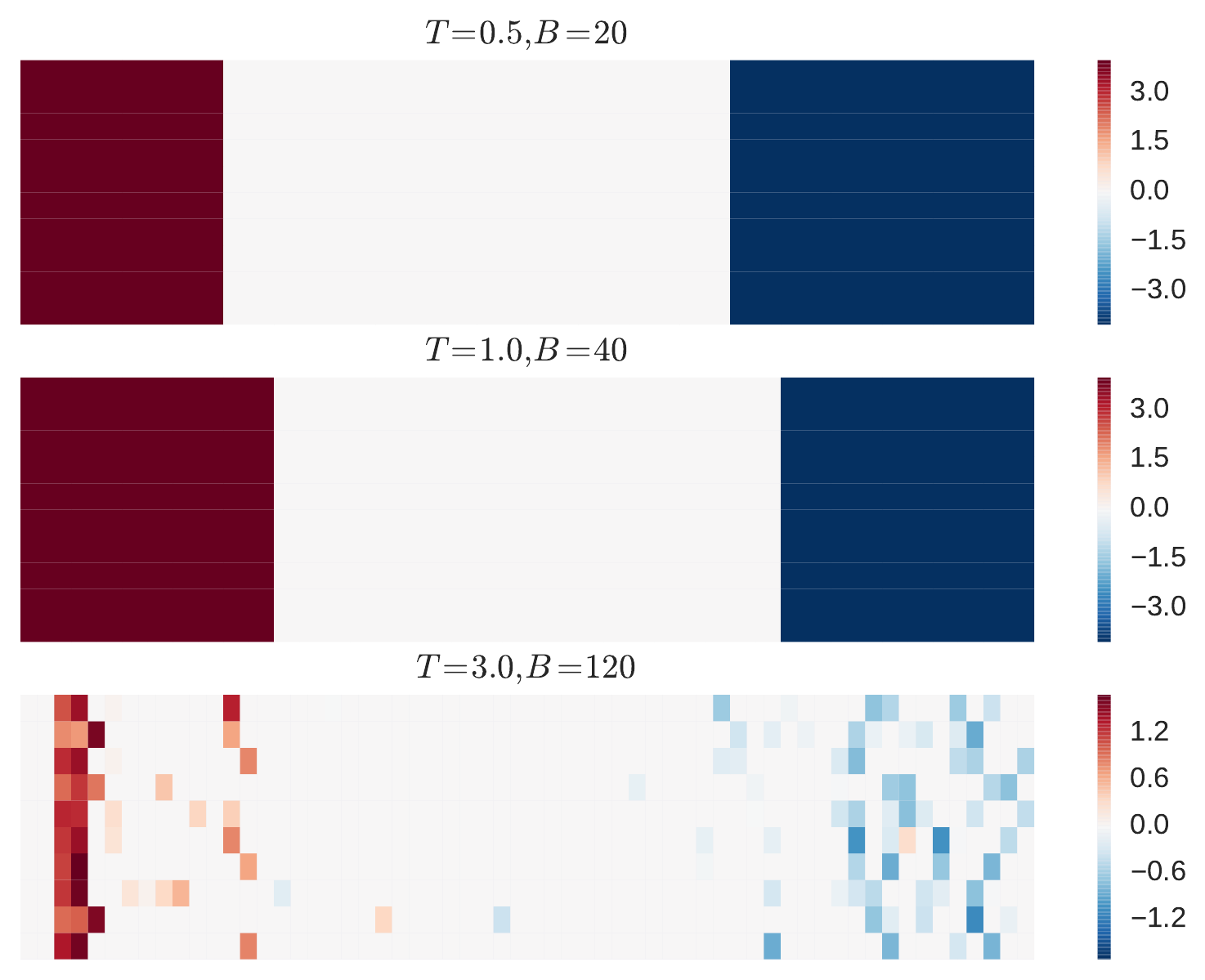}
	\caption{The 10 best sequences found for different values of $T$ and a maximal field strength $B$ amounting to half of the maximally possible. While the best sequences for $T=0.5$ and $T=1.0$ are very similar und use the maximal possible absolute field strength, the best sequences for $T=3.0$ use much smaller pulses.}
\label{fig:gs_const_seqs}
\end{figure}

\begin{figure}[t]
	\centering
	\includegraphics[width=0.4\textwidth]{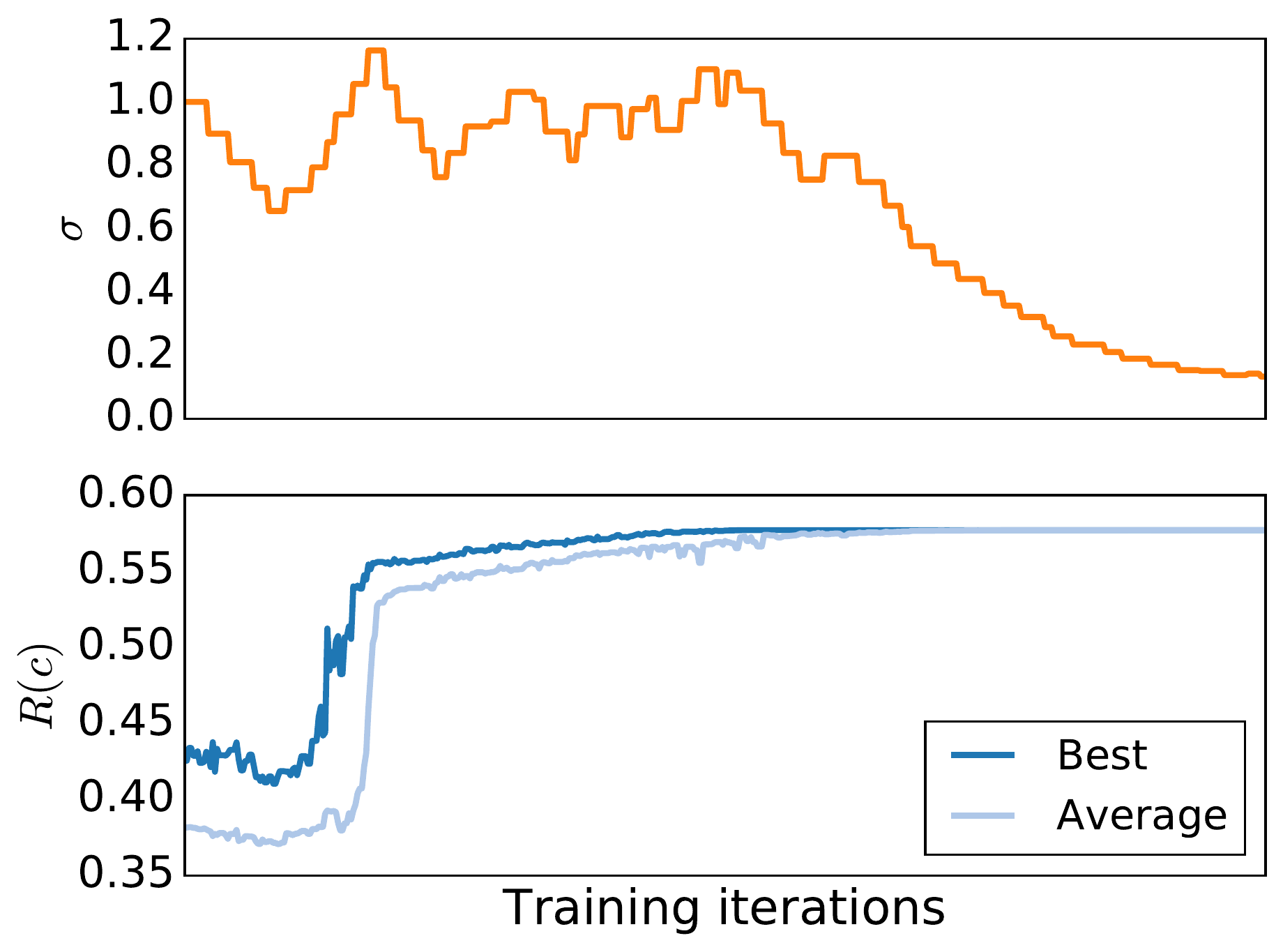}
	\caption{The convergence the best and average reward per iteration together with the dynamically adapted $\sigma$ for the constrained scenario with $T=1.0$ and $B=60$.}
\label{fig:sigma_conv}
\end{figure}

\subsection{Ground state transition}
In the ground state transition setting, we evaluate our method for times $T \in \{0.5, 1, 3\}$ with $\Delta t=0.05$ and an initial $h_i=-2$, target $h^*=2$ as well as $|h_{max}=4|$ to achieve comparability with the baseline results~\cite{bukov2017machine}. For the discrete and continuous case, we consider systems of size $L=1$ and $L=5$ and $B \in \{20,30,40,50,60,100,120,140\}$ with $L=1$ for the constrained case. Since the overlap $S_2$ as defined above already lies in the interval $[0,1]$, we used it directly as reward function, again shifting it such that a uniformly random policy achieved zero reward.

The probabilistic modelling of the sequences is similar to the quantum memory case in that we use a categorical distribution for the discrete case and a mixture-of-Gaussians for both the continuous and constrained tasks. Thereby, we model the probability density of the deviations $\Delta h_t$ in the continuous case and the predicted absolute value of $h_t$ in the constrained case.
The distributions are parameterized in the same way as above, namely by a two layer LSTM form whose output state both the discrete probabilities and the means for both discrete cases as predicted. The optimization is conducted as in the quantum memory scenario.

The results of our numerical experiments are listed in Table~\ref{tab_gs_0}. 
As shown, our method was able to replicate the baseline results from~\cite{bukov2017machine} both for the discrete and the continuous formulation of the problem for a system size of $L=1$ and also performs well for larger systems of $L=5$ with both versions yielding generally the same results. 
We indeed found the continuous version to converge to predicting zero deviation as it was expected to. 
For the constrained case we can see that our method converges to sequences whose performance is surprisingly close to the baseline results even when allowed to use only half of the maximal absolute field strength. 
For $T=3.0$ the imposed constraints in fact seem to have no negative effect as apparently already sequences with a very small total field strength suffice to achieve perfect overlap.
This is also illustraed by Figure~\ref{fig:gs_const_seqs} which shows the best 10 sequences found during the training process for $T \in \{0.5,1.0,3.0\}$ and $B$ set to half the maximal total field strength.
While for the smaller two total times the sequences are very similar and always make use of the maximal field strength or apply no pulse at all, for $T=3.0$ only the general scheme of applying positive pulses first, then doing nothing and finally applying negativ pulses persists.
The individual pulses that are applied are very weak and and entire sequence typically only amounts to a total absolute strength of $\sim 6$.
This phenomenon is likely caused by the fact that the optimization problem in this case becomes significantly easier for longer times~\cite{bukov2017machine}.
In Figure~\ref{fig:sigma_conv} we display the convergence of the best and average results sampled per iteration together with the dynamic schedule for sigma during the optimization.
It can be seen that $\sigma$ is dynamically increased when the convergence slows down, decreased when it speeds up and finally converges to a stable value as the optimization converges as well.
In other scenarios we also observed our adaption scheme to perform similarly to a decayed annealing schedule.

\section{Conclusion and Future Work}
\label{conclusion}
In this work we have tried to introduce quantum physics and especially problems in (black-box or model-free) quantum information and quantum control to a broader audience in the machine learning community and showed how they can be successfully tackled with state-of-the-art reinforcement learning methods.
To this end, we have given a brief introduction to quantum control and discussed different aspects of the application of reinforcement learning to it. 
We have argued that LSTMs are a good choice to model the sequences of control parameters arising in quantum control and shown how black-box quantum control gives rise to a particular reinforcement learning problem for whose optimization policy gradient methods are a natural choice.
As a recent and successful variant of policy gradient algorithms, we have adapted the PPO scheme for our application and introduced the MPPO algorithm.
We then went on to show how our general method for treating black-box quantum control can be easily combined with physical prior knowledge for two example scenarios and presented numerical results for a range of learning tasks arising in this context.
These results showed that our method is able to achieve state-of-the-art performance in different tasks while being able to address problems of discrete and continuous control alike and provided evidence for the hypotheses that machine learning is a good choice for the automated optimization of parameters in experiments.

This work can also be understood to some extent as a contribution to the debate about how much prior knowledge is necessary for machine learning algorithms to perform well in real-world tasks. 
During the course of this work, we have found it a necessary precondition for the addressed problems in continuous domains to be solvable to incorporate physical domain knowledge such as known good rotation axes and angles.
Without this information a reinforcement learning agent would be required to at least implicitly learn about certain laws of physics to not be lost in the infinite action space of which only a negligibly small part results in good solutions. 
This clearly is out of scope for current models and algorithms without symbolic reasoning capacity and might remain so for some time especially when the data collected by the agent is very small compared to the search space.

Finally, interesting directions of future work would be to apply the method to a real experiment and evaluate its performance there as well as to develop a set of benchmark problems in quantum control to compare the different already existing algorithms on neutral grounds.
It would also be interesting to investigate which other problems of relevance yield  reinforcement learning problems similarly structured to the formulation presented in this work.

\nocite{*}
\bibliography{ref}
\end{document}